# Planarian Neural Networks: Evolutionary Patterns from Basic Bilateria Shaping Modern Artificial Neural Network Architectures

Ziyuan Huang, Mark Newman, Maria Vaida, Srikar Bellur, Roozbeh Sadeghian, Andrew Siu, Hui Wang, and Kevin Huggins

***Abstract*—This study examined the viability of enhancing the prediction accuracy of artificial neural networks (ANNs) in image classification tasks by developing ANNs with evolution patterns similar to those of biological neural networks. ResNet is a widely used family of neural networks with both deep and wide variants; therefore, it was selected as the base model for our investigation. The aim of this study is to improve the image classification performance of ANNs via a novel approach inspired by the biological nervous system architecture of planarians, which comprises a brain and two nerve cords. We believe that the unique neural architecture of planarians offers valuable insights into the performance enhancement of ANNs. The proposed planarian neural architecture-based neural network was evaluated on the CIFAR-10 and CIFAR-100 datasets. Our results indicate that the proposed method exhibits higher prediction accuracy than the baseline neural network models in image classification tasks. These findings demonstrate the significant potential of biologically inspired neural network architectures in improving the performance of ANNs in a wide range of applications.**

***Index Terms*— biologically inspired computing, cross-network communication, dual network architecture, evolutionary neural architectures, plana[1]rian neural networks.**

## I. INTRODUCTION

THE use of multistep or modular neural networks to solve complex problems is widespread in several scientific fields, e.g., biomedical science, geography, and railroad track management [1–3]. Multistep solutions to a given research problem involve implementing and linking multiple neural networks sequentially in a series of steps and decomposing the overall task into constituent subtasks to be solved via different networks. However, when such architectures do not include cross-network communication, forward and backward propagation is possible only within each individual neural network.

In this study, we considered a planarian neural network (PNN) framework based on the biological nervous system of planarians. A PNN comprises a brain, two nerve cords, and the communication mechanisms between them. To optimize the training process, we introduced the concept of a patience gate that regulates the frequency of communication between the brain and the nerve cords. Our experiments revealed a correlation between the value of the patience gate in the PNN framework and the test accuracy, indicating that the patience gate plays a crucial role in optimizing PNN training. The best-performing PNNs on CIFAR-10 and CIFAR-100 were those with patience gate values of 15 (PNN15) and 20 (PNN20), respectively.

We adopted a standardized methodology to evaluate the performance of the proposed model on CIFAR-10 and CIFAR-100. In particular, we recorded the mean and median measurements over five runs for CIFAR-10 and seven runs for CIFAR-100. The performances of pairs of models and multiple models were compared via the Mann–Whitney U test and the Kruskal–Wallis H test, respectively.

*Corresponding author: Kevin Huggins. Harrisburg University of Science and Technology*, 326 Market Street, Harrisburg, PA 17101. Email: khuggins@harrisburgu.edu

*Ziyuan Huang is with Harrisburg University of Science and Technology, Harrisburg, Pennsylvania USA and University of Massachusetts Chan Medical School*, Worcester, Massachusetts, USA (e-mail: zhuang7@harrisburgu.edu; ziyuan.huang2@umassmed.edu).

*Mark Newman is with DevIS LLC*, Arlington, VA USA (email: mnewman@devis.com)

*Maria Vaida, Srikar Bellur, Roozbeh Sadeghian, and Kevin Huggins are with Harrisburg University of Science and Technology*, Harrisburg, Pennsylvania USA (e-mails: mvaida@harrisburgu.edu; sbellur@harrisburgu.edu; rsadeghian@harrisburgu.edu; khuggins@harrisburgu.edu)

*Andrew Siu is with Amgen Inc., Thousand Oaks, CA, USA (E-Mail: andrewjsiu@gmail.com)*

*Hui Wang is with Nanjing Audit University, Nanjing, Jiangsu, China (E-Mail: huiwang@nau.edu.cn)*

## II. MATERIALS AND METHODS

### *A. Datasets*

**Data Augmentation:** Data augmentation was performed by adding four pixels to each side of each image with values of 0 and then randomly cropping it to a 32 × 32 size, including a horizontal flip and pixel mean subtraction, during training [4–7]. During validation, only a horizontal flip was implemented. Therefore, no data augmentation was performed on the test data. This strategy was adopted to ensure that the proposed PNN algorithm can enhance model performance on the basis of only moderate data augmentation.

**CIFAR-10:** CIFAR-10 is a public dataset intended for computer vision and machine learning tasks [8]. It contains 60,000 32 × 32 color images categorized uniformly into 10 classes. These images were divided into 50,000 training images and 10,000 test images, with 5,000 and 1,000 corresponding to training and testing in each class, respectively. In this study, 10% of the training images were used as validation images—the validation data loader randomly loaded 10% of the training images in each epoch. Testing was not initiated until the training was complete.

**CIFAR-100:** CIFAR-100 is derived from the same source as CIFAR-10 [8]. CIFAR-100 contains 60,000 32 × 32 color images classified into 100 classes, with 600 images in each class. These images were categorized into 50,000 training and 10,000 test images, with 500 and 100 images belonging to each class, respectively. During k-fold cross-validation, 10% of the training images were used as validation data. Testing was performed only after the training was completed.

**Proposed PNN Architecture:** This section presents the proposed dual-nerve cord PNN adapted from the structure of the planarian nervous system. It is based on the guiding principle that the expansion, integration, and fusion of nerve integration centers facilitate nervous system evolution [9]. The PNN was designed to improve the prediction performance of artificial neural networks (ANNs). The number of artificial nerve cords was varied depending on the requirements of each experiment.

A conceptual diagram of the PNN model, comprising an artificial brain and two or more artificial nerve cords, is depicted in **Fig. 1**. The artificial brain is considered an artificial apical nervous system (AANS) that controls global activities. The nerve cords are represented by an artificial blastoporal nervous system (ABNS), which controls local and specialized functions. The artificial brain balanced global optimization on the basis of knowledge learned from the ABNS. The ABNS comprises multiple nerve cords that follow the principle of nervous system specialization, i.e., each nerve cord serves a specific purpose.

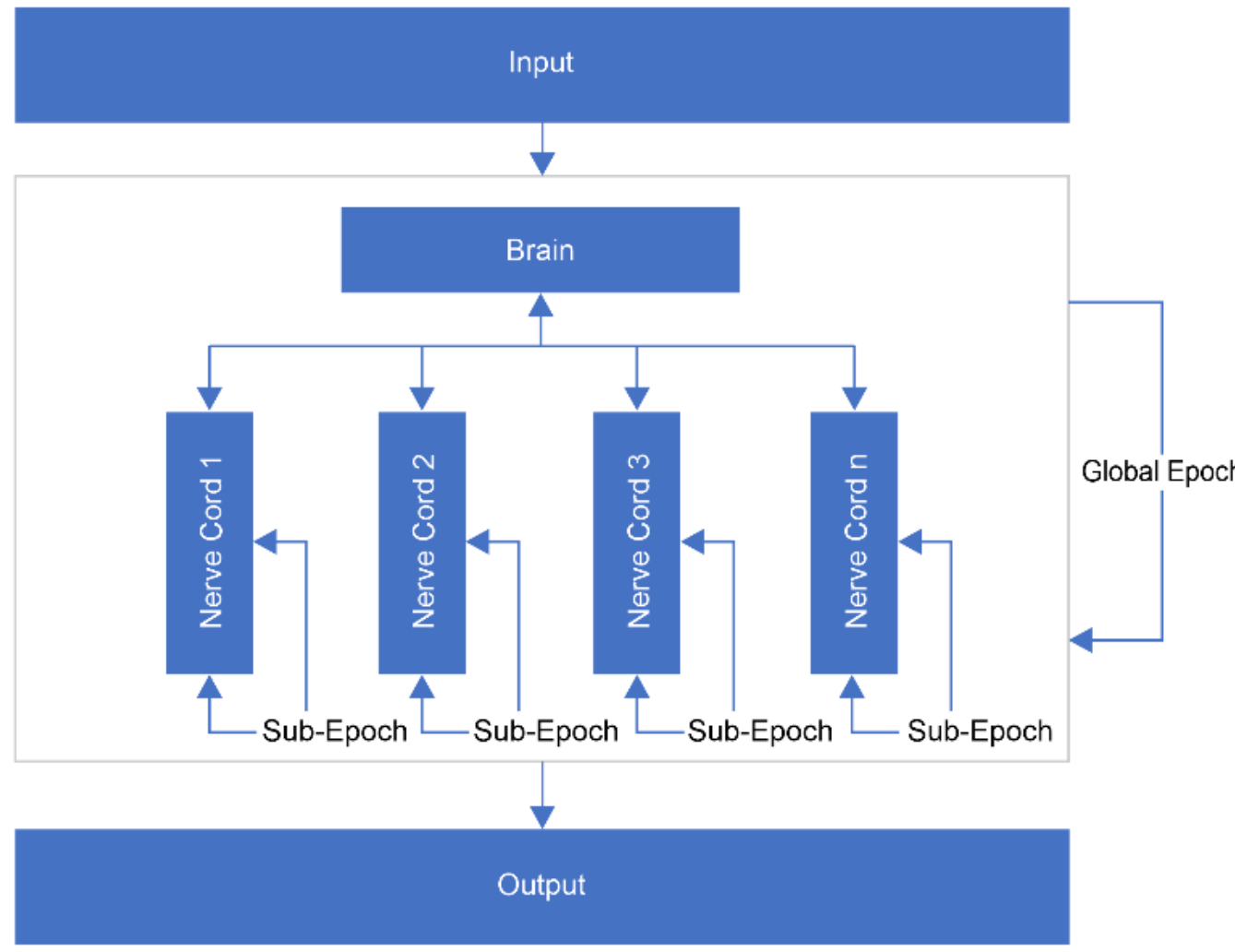

**Fig. 1**. PNN Architecture.

The architecture of the proposed PNN comprises input, nerve cord, brain, and output modules. The input module is a data loader that loads and consumes the datasets. The nerve cord module contains two or more functional neural networks. Each nerve cord performs specific functions. In the brain module, nerve cords exchange information and balance results. Finally, the output module presents the results.

### *B. Experimental Design*

We conducted experiments on the CIFAR datasets using Python 3.8.11 and PyTorch 1.12.0 on a Ubuntu 20.04 workstation with four RTX 3090 graphics cards. The proposed PNN framework was evaluated using two sets of experiments—on CIFAR-10 and CIFAR-100. Each experimental set included 12 models, with five random seeds for CIFAR-10 and seven random seeds for CIFAR-100. Each experiment corresponded to the execution of one model with one seed, and the mean result of each model over the different seeds was considered to be its final result on each dataset.

Through systematic random sampling, we selected five and seven initial seeds from the seed lists for CIFAR-10 and CIFAR-100, respectively. Systematic random sampling was employed owing to two advantages: 1) it covers the population more evenly, and 2) it is easier to perform than other alternatives [10]. For the seed population, we used the lower and upper value boundaries of NumPy's seed generator, which ranged between 0 and 2**32–1. The interval, k, was calculated by dividing 2**32–1 by 60. Each $k^{th}$ integer represented the expected seed number for our experiments. On CIFAR-10, five of the 60 seeds were used for each model, as both the first and second ResNet publications reported model performance in terms of five-run averages [4,5]. For CIFAR-100, seven seeds were used to increase the precision of the experimental results, as CIFAR-100 is considered to be significantly more complex than CIFAR0-10.

## III. Model Selection and Architecture Construction

At the time of writing, ResNet is one of the most well-established deep neural network families and provides a systematic and robust development foundation for both wide and deep network architectures compared with other deep architectures for learning algorithm families. Consequently, we considered ResNet and its variants as reference deep learning algorithms in this study, inspired by a study conducted by Google Research on deep and wide neural networks [11]. According to Eguyen *et al.*, deep networks perform better on consumer goods, whereas wide networks perform better on classes that represent scenes. Nevertheless, both network types achieved similar levels of accuracy in our study.

In this study, we constructed deep and wide neural networks with similar numbers of neurons to evaluate the performance of the proposed models and demonstrated that the improvements in test accuracy were induced by the inclusion of cross-network communication rather than an increase in the number of neurons, which is already known to improve neural network performance [12–15].

### *A. Brain, Nerve Cord, and Interaction between Them*

The brain and nerve cords are the most critical components of the proposed PNN. Our experiments involved an artificial brain that enabled nerve cords to exchange weights during training. Most modern ANNs, e.g., ResNet, DenseNet, VGG, Inception, and PeleeNet, implement StemBlocks initially. The PeleeNet study specifically discussed StemBlocks, which enhance the feature expression abilities of a neural network without significantly increasing the computational burden [16]. StemBlocks are usually independent of complicated ConvBlocks in the aforementioned neural networks and have a relatively simple architecture. These characteristics make the StemBlock an ideal weight-exchange portal between the brain and nerve cords in the PNN architecture.

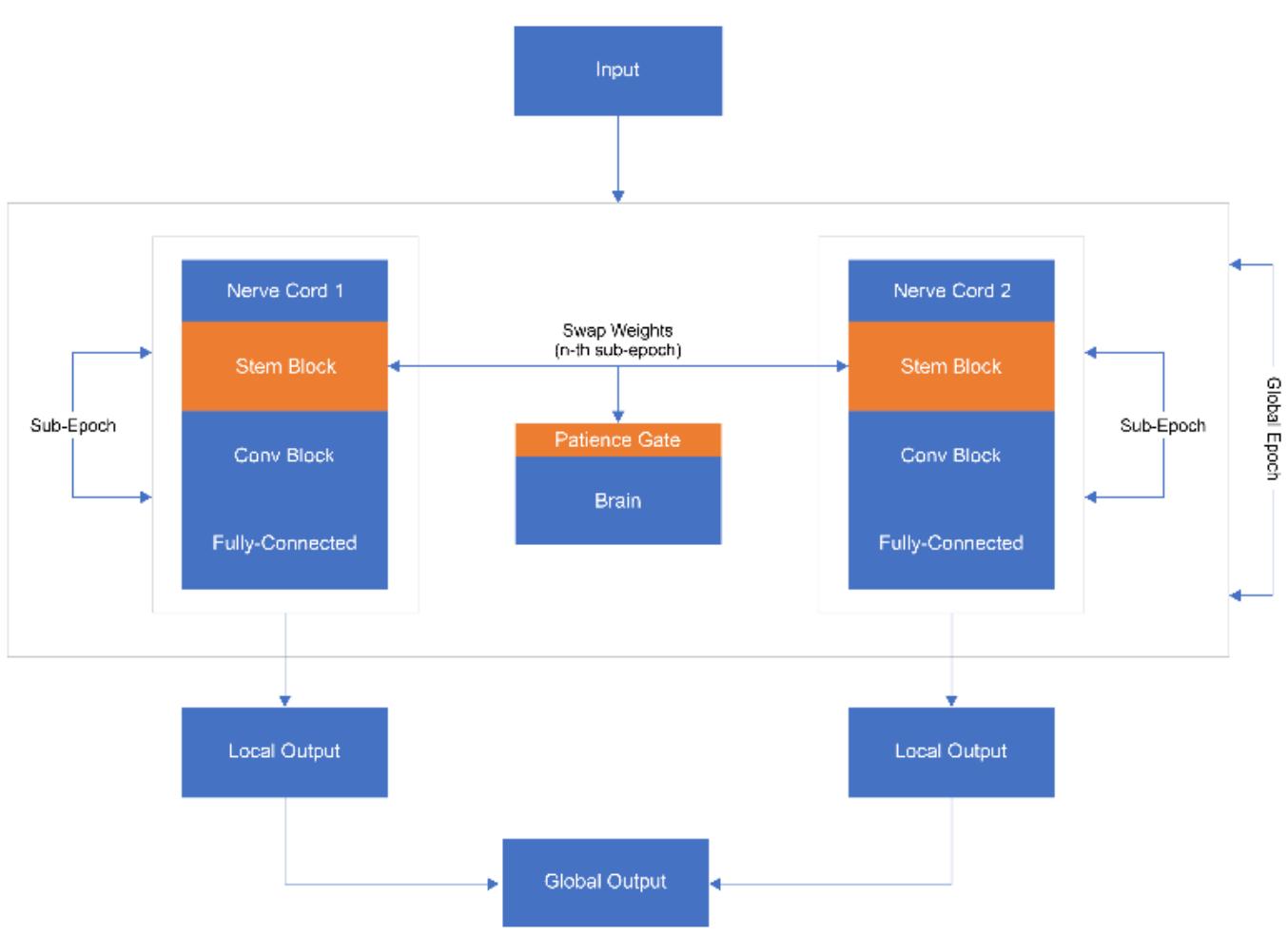

**Fig. 2.** StemBlock-based weight exchange in PNN.

In the proposed PNN, the two nerve cords were trained over subepochs, and their learned knowledge was balanced over global epochs in the artificial brain (as illustrated in **Fig. 2**. and Algorithm 1 in **TABLE I**.). During training, the nerve cords and the brain interacted by exchanging information via StemBlocks. The number of subepochs allocated to each nerve cord determined the interaction time during the global epoch. At the end of each subepoch, the weights of the StemBlocks corresponding to both nerve cords were transferred to the brain and interchanged. The frequency of weight swapping was determined by a parameter, n, indicating that weights were interchanged once every $n^{th}$ subepoch. The following pseudocode demonstrates the interaction of nerve cords with the brain via StemBlocks—a global epoch was taken to comprise one fully executed subepoch for each nerve cord.

TABLE I
WEIGHT EXCHANGE PSEUDOCODE, PYTORCH-LIKE, OF THE PNN

**Algorithm 1**

```
for epoch in range(global_epochs):

    # When the weight swap condition is equal to True
    # This is the artificial brain where nerve cord's weight exchanges cross-network
    # See Algorithm 2 for details
    if weight_swap_condition == True:
            brain(weight_swap_function,
                    net_1.stemblock(), net_2.stemblock(), …, net_n.stemblock())

    # This is where the nerve cords are trained
    for sub_epoch in range(sub_epochs_1):
            train(net_1, train_data, validation_data)

    for sub_epoch in range(sub_epochs_2):
            train(net_2, train_data, validation_data)
    …
    …
    for sub_epoch in range(sub_epochs_n):
            train(net_n, train_data, validation_data)

# Individual nerve cord generates outputs
output_1 = predict(net_1, test_data)
output_2 = predict(net_2, test_data)
...
output_n = predict(net_n, test_data)

# Ensemble technique to summarize the results as a single output
final_output = ensemble_classifier(output_1, output_2, …, output_n)
```

### *B. Weight Swap Condition*

The weight-swapping mechanism was managed using a patience gate, which is a crucial concept in the PNN framework (**Fig. 2**.). Better optimization was achieved by controlling weight swapping during training. Algorithm 2 (**TABLE II**.) describes the interaction of the weight-swapping conditions with the brain and nerve cords.

TABLE II
PATIENCE GATE PSEUDOCODE, PYTORCH-LIKE

**Algorithm 2**

```
Set patience_level = 0
Set max_patience = n

for epoch in range(global_epochs):

    if current_best_acc > new_generated_acc:
        patience_level+=1
        weight_swap_condition = (patience_level>max_patience)
    else:
        patience_level=1
        weight_swap_condition=False

    if weight_swap_condition == True:
        brain(weight_swap_function,
                net_1.stemblock(), net_2.stemblock(), …, net_n.stemblock())
        # reset the patience gate and weight swap condition after each swap
        patience_level=0
        weight_swap_condition=False
```

The aggressiveness of the nerve cords was governed by swapping their weights using StemBlocks depending on the maximum value of the patience parameter. The weights were interchanged according to the following logic. When the maximum patience was set to 1, the weights were interchanged whenever the newly calculated validation accuracy was lower than the best validation accuracy observed during training. When the maximum patience was set to 10, the weights were interchanged whenever the newly calculated validation accuracy was lower than the best validation accuracy observed over ten consecutive epochs. If the new validation accuracy was greater than the previous best validation accuracy, then maximum patience was attained. In this case, the patience level was reset to 1, and no weights were interchanged. Therefore, each weight swap reset the patience level to 1.

### *C. Models used in Experiments on CIFAR-10*

ResNet20, WideResNet14, Ensemble (ResNet20 + WideResNet14), and a PNN (ResNet20 + WideResNet14) were designed, proposed, trained, and evaluated on CIFAR-10. The loss function, optimizer, and scheduler used were cross-entropy, stochastic gradient descent, and cosine annealing, respectively. The weight decay, momentum, initial learning rate, and batch size were 0.0001, 0.9, 0.1, and 128, respectively. Two hundred epochs were considered instead of 64,000 iterations, as reported in the first and second ResNet publications. Among the aforementioned models, ResNet20, WideResNet14, and Ensemble (ResNet20 + WideResNet14) are non-PNN models, whereas PNN (ResNet20 + WideResNet14) is selected as the PNN model. The location of the network within the PNN framework did not affect its performance. The models are described below:

**ResNet20**: ResNet20 was originally introduced by He *et al*. [4] in their first ResNet study using the CIFAR-10 dataset. As a baseline, we replicated the ResNet20 model and its results on the basis of the description in the first ResNet paper [4]. The constructed ResNet20 consisted of 6n+2 weighted layers, where n = {3}. Each n represents three basic blocks, each containing two convolution layers. The first convolution layer comprised a $3 \times 3$ convolution with a $1 \times 1$ stride and $1 \times 1$ padding. A stack of 6n $3 \times 3$ convolution layers was subsequently constructed with feature maps of sizes 32, 16, and 8. Each feature map size corresponded to 2n layers. Therefore, 16, 32, and 64 filters were used, respectively. Subsampling in ResNet20 used a stride size of $2 \times 2$ pixels. Subsequently, global average pooling was implemented at the end of the neural network with a 10-way fully connected layer. The total number of neurons in ResNet20 was 272,474.

**WideResNet14:** We adopted the concept of wide neural networks in this study in the form of wide residual networks (WRNs) [17]. On the basis of WRNs and ResNet20, we developed WideResNet14, which comprises 4n+2 weighted layers, where n = {3}. Each n represents two basic blocks, each containing two convolution layers. The number of filters was set to 32 and 64 to widen the network. The other configurations of WideResNet14 were identical to those of ResNet20. Consequently, the total number of neurons in WideResNet14 was 258,458.

**Ensemble (ResNet20 + WideResNet14):** An ensemble-stacking technique was implemented in the ensemble experiment (**Fig. 3**.). ResNet20 and WideResNet14 were allowed to operate independently. The two models were stacked in parallel, and a Softmax classifier was used to generate the results. Finally, using soft voting with prediction probabilities, the two sets of results were merged and integrated into one set of predictions. The Ensemble model included 530,932 neurons in aggregate.

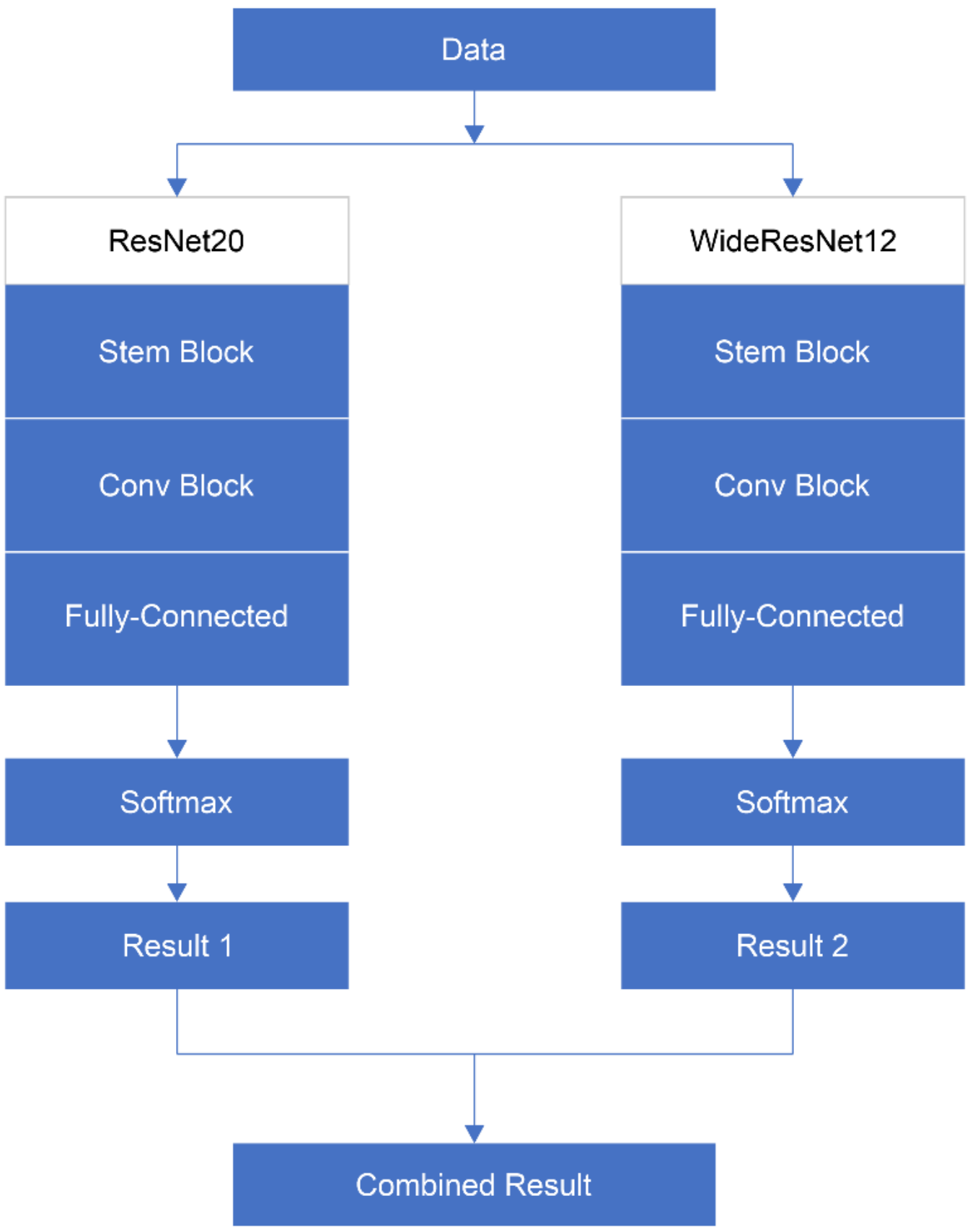

**Fig. 3.** Ensemble (ResNet + WideResNet).

**PNN (ResNet20 + WideResNet14):** PNN (ResNet20 + WideResNet14) was implemented via ensemble learning (**Fig. 3**.) and two nerve cords with brain interactions (**Fig. 2**.). Two types of epochs were considered in the PNN model—global epochs and subepochs. A global epoch was considered when all nerve cords completed a set of forward and backward propagations. A subepoch was considered when one nerve cord completed a set of forward and backward propagations using the entire training dataset. Moreover, a weight swap referred to the transfer of weights from one nerve cord stem block to another. In this study, the learned knowledge leaked from one neural network to another during the weight-swapping process, enabling the two neural networks within the PNN framework to learn from each other. The PNN model contained 530,932 neurons in aggregate.

*D. Models used in Experiments on CIFAR-100*

Four types of networks were used on CIFAR-100—ResNet164, WideResNet110, Ensemble (ResNet164 + WideResNet110), and PNN (ResNet164 + WideResNet110). The hyperparameters used in these experiments were identical to those used in the case of CIFAR-10. ResNet164, WideResNet110, and Ensemble (ResNet164 + WideResNet110) were independently used for non-PNN models, whereas the stacked ResNet164 + WideResNet110 combination was considered in the case of PNN to be examined, compared, and analyzed. The first nerve cord in the PNN model was ResNet164, and the second was WideResNet110. The locations of the nerve cords were not significant.

**ResNet164:** We replicated ResNet164 [4,5] based on the first and second ResNet studies. Its model architecture was nearly identical to that of ResNet20—the only exception was that n = 18 with three bottleneck blocks and the number of classes was changed to 100. Owing to the inclusion of three convolution layers in each bottleneck block, ResNet164 contained 9n+2 weighted layers with 1,727,284 neurons.

**WideResNet110:** We constructed WideResNet110 based on design concepts presented in the first and second ResNet studies and the WideResNet study [4,5,17]. WideResNet110 contained six 6n+2 weighted layers, with n = {18}. Each n represents three bottleneck blocks, and each bottleneck block contained two convolutional layers. In addition, the number of filters was set to 32 and 64 to widen the network. The other configurations of WideResNet110 were identical to those of WideResNet14. In aggregate, WideResNet110 contained 1,637,428 neurons.

**Ensemble (ResNet164 + WideResNet110):** We constructed the Ensemble (ResNet164 + WideResNet110) network using the neural network architectural methodologies of ResNet, WideResNet, and ensemble learning. The ensemble models, ResNet164 and WideResNet110, were trained independently. After training and validation, ResNet164 and WideResNet110 were stacked as ensemble models. The soft voting technique of Ensemble (ResNet164 + WideResNet110) is identical to that of Ensemble (ResNet20 + WideResNet14). On the basis of the prediction probabilities, Ensemble (ResNet164 + WideResNet110) produced one set of predictions. This Ensemble model comprised 3,364,712 neurons.

**PNN (ResNet164 + WideResNet110):** PNN (ResNet164 + WideResNet108) was constructed using a similar configuration as PNN (ResNet20 + WideResNet12). It exhibited identical global epochs, subepochs, and weight-swapping principles but was deeper than the latter. ResNet164 and WideResNet108 were stacked together as two nerve cords with brain interactions using StemBlocks. Each nerve cord generated its own results, and soft voting was used to merge the two sets of predicted results into one. The PNN model contained 3,364,712 neurons.

## IV. Results

For CIFAR-10 and CIFAR-100, five and seven seeds were used per model, respectively, and one seed was used per run. The models were evaluated via methods discussed in the first and second ResNet publications, with modifications presented in this section [4,5]. For CIFAR-10 and CIFAR-100, He *et al.* [4,5] suggested that the mean and median over five runs, respectively, adequately represented model performance. In this study, the number of CIFAR-100 runs per model was increased to seven to increase the precision of the results. The results obtained on CIFAR-10 and CIFAR-100 are presented in Tables III, IV, V, and VI.

### *A. Experimental Results on CIFAR-10*

The experimental results on the CIFAR-10 dataset are presented in Tables III and IV. These tables list the neural networks and the number of networks used in each model, indicating that each model was trained and tested using five different seeds. Table III lists the test error rates of single networks with five seeds, where each model comprises only one neural network. Table IV lists the test error rates of dual networks with five seeds, where each model comprises two neural networks.

**Single Network Results:** Eight models were used in a single network. The baseline models were ResNet20 and WideResNet14. The remaining six models were trained via the PNN framework, in which each PNN comprised two neural networks. Each PNN was considered in terms of two independent models to observe the effect of the PNN architecture on the prediction performance of individual neural networks. On CIFAR-10, after the PNN models were fine-tuned, we presented the PNN5, PNN10, and PNN15 models, each containing ResNet20 and WideResNet14, with patience gate values of 5, 10, and 15, respectively. The five-run averages for ResNet20, WideResNet14, PNN5 ResNet20, PNN5 WideResNet14, PNN10 ResNet20, PNN10 WideResNet14, PNN15 ResNet20, and PNN15 WideResNet14 are listed in **TABLE III**.

In **TABLE III**, the underlined numbers represent averages over the five seeds in experiments on CIFAR-10. The average error rate of PNN15's ResNet20 was observed to be 7.27, which was the best among the eight models. The average error rate of PNN15's WideResNet14 was the second best. As the patience gate value increased, the five-run error average decreased steadily.

TABLE III
ERROR RATES (%) OF SINGLE NETWORK ARCHITECTURES ON CIFAR-10

| **Error Rates (%) of Single Networks on CIFAR-10** | | | | | | | |
|---|---|---|---|---|---|---|---|
| **Network** | **Sub-Network** | Seed1 | Seed2 | Seed3 | Seed4 | Seed5 | Mean |
| **ResNet20** | - | 7.50 | 7.38 | 7.57 | 7.49 | 7.24 | 7.44 |
| **WideResNet14** | - | 7.51 | 7.66 | 7.25 | 7.48 | 7.89 | 7.56 |
| **PNN5** | **ResNet20** | 8.10 | 7.78 | 7.17 | 7.43 | 7.83 | 7.66 |
| | **WideResNet14** | 7.55 | 7.81 | 7.78 | 7.84 | 7.18 | 7.63 |
| **PNN10** | **ResNet20** | 7.88 | 7.53 | 7.49 | 7.16 | 7.35 | 7.48 |
| | **WideResNet14** | 7.29 | 7.56 | 7.41 | 7.36 | 7.91 | 7.51 |
| **PNN15** | **ResNet20** | 7.38 | 7.02 | 7.22 | 7.24 | 7.48 | **7.27** |
| | **WideResNet14** | 7.49 | 7.26 | 7.22 | 7.63 | 7.24 | **7.37** |

**Dual Network Results:** Each dual network comprised four models. The baseline model was an Ensemble (ResNet20 + WideResNet14) model. In addition to the baseline model, PNN5 (ResNet20 + WideResNet14), PNN10 (ResNet20 + WideResNet14), and PNN15 (ResNet20+WideResNet14) were included, with fine-tuned patience gate values of 5, 10, and 15, respectively. The five-run average errors of Ensemble (ResNet20 + WideResNet14), PNN5 (ResNet20 + WideResNet14), PNN10 (ResNet20 + WideResNet14), and PNN15 (ResNet20 + WideResNet14) were 5.98, 6.01, 5.94, and 5.81, respectively.

Among the four dual networks listed in **TABLE IV**, PNN15 (ResNet20 + WideResNet14) exhibited the best five-run average error of 5.81 compared with the baseline model's 5.98. We also observed a decreasing trend in the five-run average test error as the patience gate value increased.

TABLE IV
ERROR RATES (%) OF DUAL NETWORK ARCHITECTURES ON CIFAR-10

| **Error Rates (%) of Dual Network Architectures on CIFAR-10** | | | | | | | |
|---|---|---|---|---|---|---|---|
| **Network** | **Sub-Network** | Seed1 | Seed2 | Seed3 | Seed4 | Seed5 | Mean |
| **Ensemble** | **ResNet20 WideResNet14** | 6.02 | 5.87 | 6.02 | 5.88 | 6.12 | 5.98 |
| **PNN5** | **ResNet20 WideResNet14** | 5.95 | 6.07 | 5.89 | 5.87 | 6.25 | 6.01 |
| **PNN10** | **ResNet20 WideResNet14** | 5.98 | 6.05 | 5.99 | 5.71 | 5.98 | 5.94 |
| **PNN15** | **ResNet20 WideResNet14** | 5.87 | 5.80 | 5.67 | 5.82 | 5.88 | **5.81** |

### *B. Experimental Results on CIFAR-100*

The results on CIFAR-100 are presented in Tables V and VI, which list the neural networks used in each model and the number of networks involved in each model, indicating that each model was trained and tested using seven different seeds. We evaluated the models in terms of three measures—the mean, trimmed mean, and median. The trimmed mean was calculated after removing the highest and lowest values from the original test results. **TABLE V** lists the test error rates of the eight single networks for seven seeds. **TABLE VI** presents the test error rates of the four dual networks—one ensemble and three PNN methods—for seven seeds.

**Single Network Results:** Eight models were used in the single-network architecture. The two baseline models were ResNet164 and WideResNet110. Three pairs of ResNet164 and WideResNet110 were trained within the PNN framework with patience gate values of 10 (henceforth referred to as PNN10), 15 (henceforth PNN15), and 20 (henceforth PNN20), respectively. We treated each neural network embedded in the PNN framework as an independent model to observe the effect of the PNN architecture on the prediction performance of individual neural networks. **TABLE V** presents the seven-run errors for ResNet164, WideResNet110, PNN10 ResNet164, PNN10 WideResNet110, PNN15 ResNet164, PNN15 WideResNet110, PNN20 ResNet164, and PNN20 WideResNet110.

PNN15's WideResNet110 performed the best among the eight single-network models in terms of the mean, trimmed mean, and median. On CIFAR-100, fine-tuning the patience gate values was observed to improve the test error rates.

TABLE V
ERROR RATES (%) OF SINGLE NETWORK ARCHITECTURES ON CIFAR-100

**Error Rates (%) of Single Network Architectures on CIFAR-100**

| **Network** | **Sub-Network** | Seed 1 | Seed 2 | Seed 3 | Seed 4 | Seed 5 | Seed 6 | Seed 7 | Mean | Trimmed Mean | Median |
|---|---|---|---|---|---|---|---|---|---|---|---|
| **ResNet164** | - | 23.64 | 23.67 | 23.11 | 22.66 | 23.85 | 23.54 | 23.99 | 23.49 | 23.56 | 23.64 |
| **WideResNet110** | - | 23.10 | 22.39 | 25.49 | 23.90 | 22.56 | 22.76 | 25.31 | 23.64 | 23.53 | 23.10 |
| **PNN10** | **ResNet164** | 24.01 | 23.26 | 23.56 | 23.24 | 24.11 | 23.91 | 21.38 | 23.35 | 23.60 | 23.56 |
| | **WideResNet110** | 22.55 | 22.95 | 25.91 | 23.98 | 22.40 | 22.84 | 24.10 | 23.53 | 23.28 | 22.95 |
| **PNN15** | **ResNet164** | 24.59 | 23.29 | 24.04 | 23.37 | 24.31 | 23.97 | 24.00 | 23.94 | 23.94 | 24.00 |
| | **WideResNet110** | 22.09 | 22.02 | 25.54 | 23.65 | 22.52 | 22.58 | 24.23 | **23.23** | **23.01** | **22.58** |
| **PNN20** | **ResNet164** | 24.04 | 23.18 | 23.65 | 23.37 | 24.31 | 23.59 | 23.66 | 23.69 | 23.66 | 23.65 |
| | **WideResNet110** | 21.99 | 22.50 | 26.17 | 23.65 | 22.52 | 22.68 | 24.07 | 23.37 | 23.08 | 22.68 |

**Dual Network Results: TABLE VI** lists the test error rates of the four dual network models with seven seeds on CIFAR-100. The baseline model was an ensemble (ResNet164+WideResNet110), and the proposed models were PNN10 (ResNet164+WideResNet110), PNN15 (ResNet164+WideResNet110), and PNN20 (ResNet164+WideResNet110). The PNN models were fine-tuned using patient gate values of 10 (PNN10), 15 (PNN15), and 20 (PNN20).

The seven-run average errors of the Ensemble (ResNet164+WideResNet110), PNN10 (ResNet164 + WideResNet110), PNN15 (ResNet164 + WideResNet110), and PNN20 (ResNet164+WideResNet110) models were 20.79, 21.36, 20.82, and 20.67, respectively. Their seven-run trimmed mean values were 20.71, 20.92, 20.78, and 20.55, respectively, and their seven-run medians were 20.67, 20.77, 20.64, and 20.46, respectively. In terms of these metrics, PNN20 (ResNet164 + WideResNet110) performed the best among the four models. The results presented in **TABLE VI** demonstrate that fine-tuning the PNN framework reduced test error rates, which correlated with an increase in the patient gate value.

TABLE VI
ERROR RATES (%) OF DUAL NETWORK ARCHITECTURE ON CIFAR-100

**Error Rates (%) of Dual Network Architecture on CIFAR-100**

| Network | Sub-Network | Seed 1 | Seed 2 | Seed 3 | Seed 4 | Seed 5 | Seed 6 | Seed 7 | Mean | Trimmed Mean | Median |
|---|---|---|---|---|---|---|---|---|---|---|---|
| **Ensemble** | **ResNet164 WideResNet110** | 20.77 | 20.27 | 21.08 | 20.45 | 20.67 | 20.59 | 21.72 | 20.79 | 20.71 | 20.67 |
| **PNN10** | **ResNet164 WideResNet110** | 20.62 | 20.77 | 21.53 | 20.75 | 20.53 | 20.92 | 24.40 | 21.36 | 20.92 | 20.77 |
| **PNN15** | **ResNet164 WideResNet110** | 20.56 | 20.27 | 21.58 | 20.80 | 20.46 | 20.62 | 21.48 | 20.82 | 20.78 | 20.62 |
| **PNN20** | **ResNet164 WideResNet110** | 20.27 | 20.21 | 21.76 | 20.80 | 20.46 | 20.41 | 20.79 | **20.67** | **20.55** | **20.46** |

## V. Discussion

The experimental results obtained on the CIFAR-10 and CIFAR-100 datasets were analyzed via the nonparametric Kruskal–Wallis H test (h test) to define statistically significant differences between multiple models and the Mann–Whitney U test (u test) to determine statistically significant differences between pairs of models. Owing to greater data complexity, the results corresponding to CIFAR-100 were more likely to not be normally distributed. The prediction performances on CIFAR-10 and CIFAR-100 were measured in terms of five-run medians, as in the second ResNet publication [5]. In this study, we utilized nonparametric tests of medians, which were deemed appropriate for this type of data.

The H test assumes ordinal or continuous variables, independent samples, and sufficient data [18]. Assumption 1 required ordinal or continuous observations of hierarchical relationships. This was satisfied—in experiments on both CIFAR-10 and CIFAR-100, the test errors were continuous observations with hierarchical relationships. In this hierarchy, lower outputs corresponded to higher quality. Assumption 2 requires that observations be independent samples. This was also satisfied as the sample data were derived from the models introduced in this study. Each model generated its own results independently of the others. The final assumption was the availability of sufficient data. For the experiments on CIFAR-10, five samples were generated per group, whereas for those on CIFAR-100, seven samples were generated per group.

The assumptions of the U test were that the two groups must be independent and the dependent variable must be ordinal or continuous [19]. These were satisfied by the experimental design on both CIFAR-10 and CIFAR-100, as each model generated its own data to be used in the U tests, thus ensuring independence. Further, the sample data generated in the experiments were continuous because they were test errors that measured model performance.

Hence, all the experimental results satisfied the assumptions of the H and U tests. The following discussion focuses on the results of visual observations, H tests, and U tests.

### *A. Discussion of Results on CIFAR-10*

**Visual Observation:** The experimental results on CIFAR-10 were plotted jointly to demonstrate the error rates of all the models, with both single and dual networks. The eight models with a single-network architecture presented a mean error rate of 7.49, whereas the four models with a dual-network architecture presented a mean error rate of 5.93. Because of the obvious mean differences between the single and dual networks described below (**Fig. 4**.), we decided to analyze the two cases separately to determine the impact of the PNN framework on model performance.

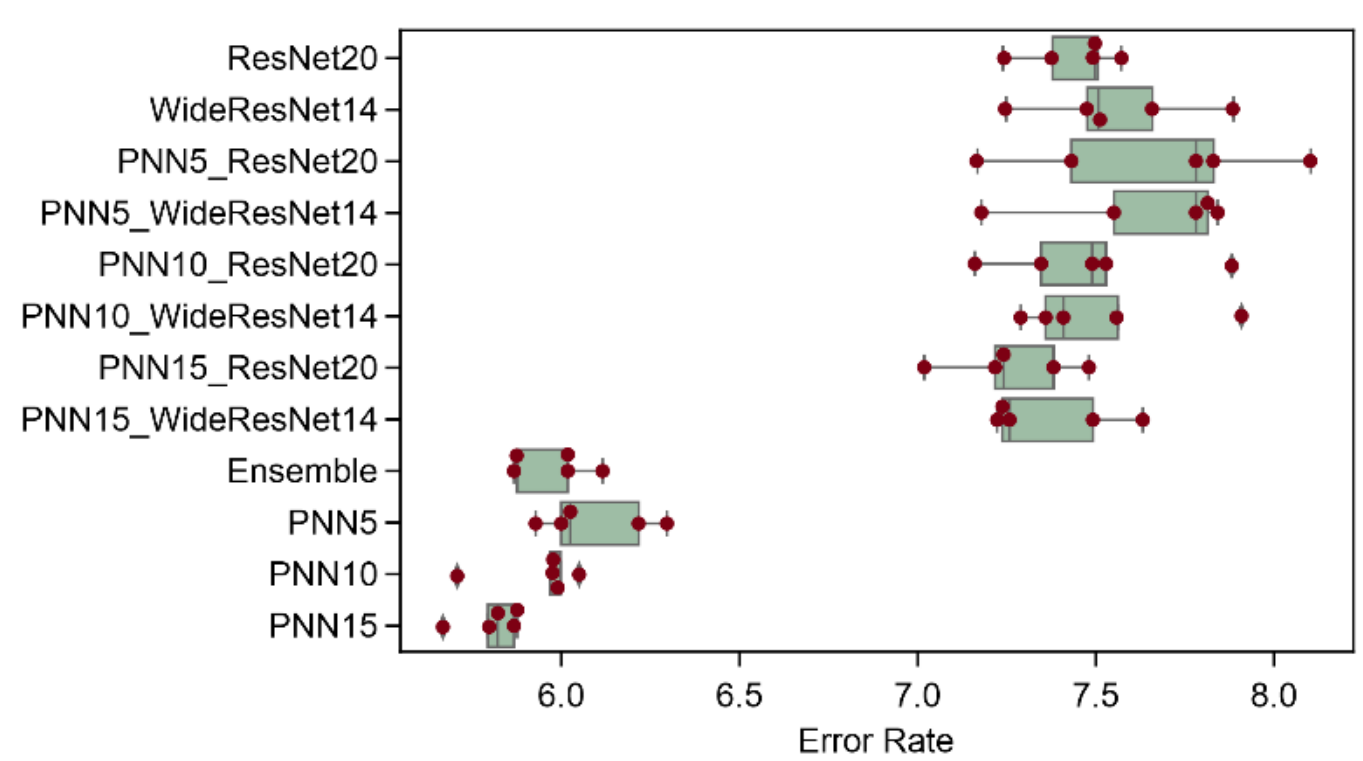

**Fig. 4.** Error rates of the models on CIFAR-10.

**Single Network Architecture:** The single-network PNN models exhibited certain advantages over the baseline models (**Fig. 5.**). To verify the visual observations, H tests and U tests were conducted.

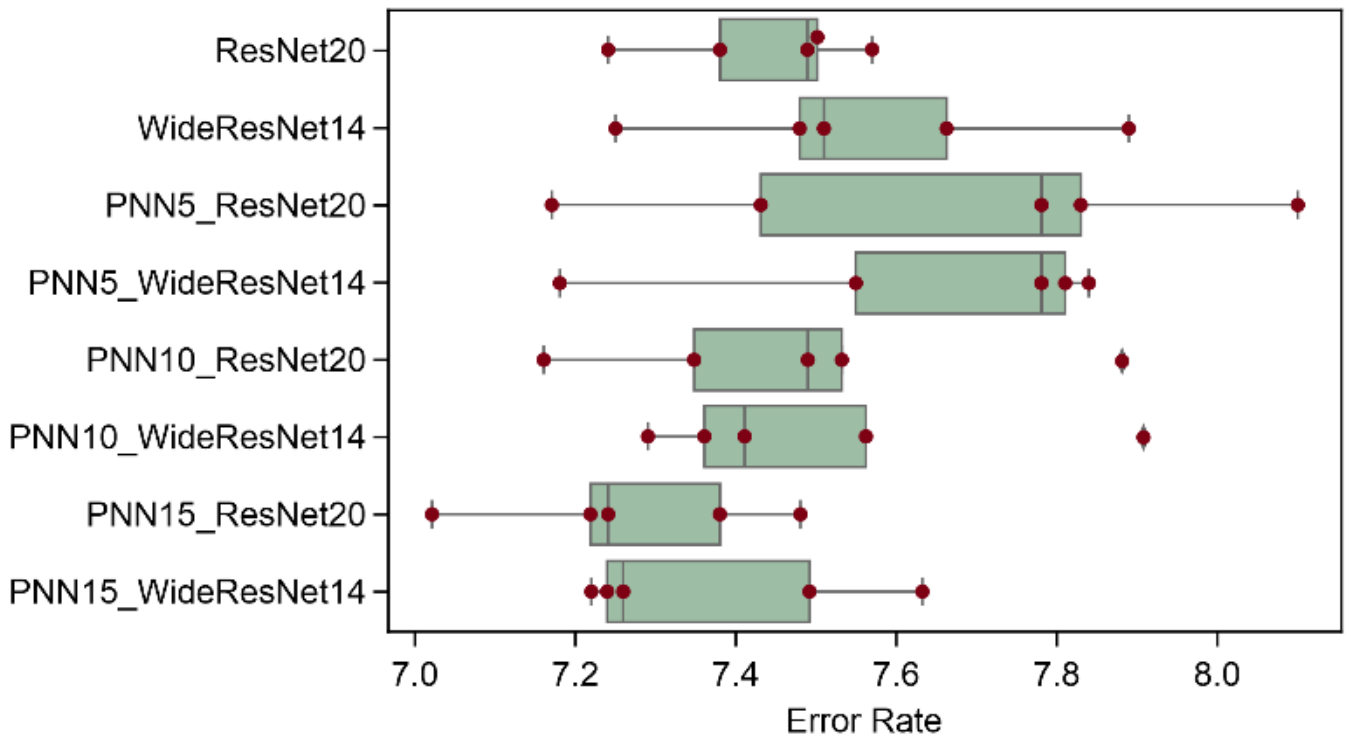

**Fig. 5.** Error rates of single-network models on CIFAR-10.

The H-test hypotheses on single network models during experiments on CIFAR-10 are listed below:

$$H_0: \tilde{x}_1 = \tilde{x}_2 = \tilde{x}_3 = \tilde{x}_4 = \tilde{x}_5 = \tilde{x}_6 = \tilde{x}_7 = \tilde{x}_8$$
$$H_1: at\ least\ one\ population\ median\ value\ is\ different\ from\ those\ of\ other\ models$$

The population medians of the eight experimental models tested on CIFAR-10 are denoted by $\tilde{x}_1, \tilde{x}_2, \tilde{x}_3, \tilde{x}_4, \tilde{x}_4, \tilde{x}_6, \tilde{x}_7,$ and $\tilde{x}_8$ in the H test null hypothesis.

The H statistic is given by:

$$H = \frac{12}{N(N+1)} \sum_{i=1}^{k} \frac{R_i^2}{n_i} - 3(N+1) \qquad (1)$$

where $N$ denotes the total number of observations, $k$ denotes the number of models or groups in the CIFAR-10 experiments, $R_i$ denotes sample $i$'s sum of ranks per group, and $n_i$ denotes the number of observations per group. Using SciPy's statistical function for the H test, the H statistic was calculated to be 8.27, with a p value of 0.31. Given α = 0.05, which was less than the p value of 0.31, there was not enough evidence to reject the null hypothesis and declare the differences between the test errors of the eight models to be significant.

However, the differences between the best-performing PNN and baseline models still had to be investigated. Hence, we performed U tests on two pairs of comparisons—ResNet20 vs. PNN15's ResNet20 and WideResNet14 vs. PNN15's WideResNet14—to evaluate whether the PNN framework improves model performance. The hypotheses of the U test are listed below, where $\tilde{x}_b$ represents the population median of the baseline models (ResNet20 or WideResNet14) and $\tilde{x}_p$ represents the population median of the proposed models (PNN15's ResNet20 or PNN15's WideResNet14). Therefore, the hypothesis testing statements were as follows:

$H_0: \tilde{x}_b = \tilde{x}_p$, two population medians are identical

$H_1: \tilde{x}_b \neq \tilde{x}_p$, two population medians are different

For the first pair of U tests, we considered a sample of $n_b$ observations {7.5, 7.38, 7.57, 7.49, 7.24} for ResNet20 and a sample of $n_p$ observations {7.38, 7.02, 7.22, 7.24, 7.48} for PNN15's ResNet20. Using SciPy's statistical function for the U test, we generated a statistic of 21.00 with a p value = 0.09. Given α = 0.05, the null hypothesis that ResNet20 and PNN15's ResNet20 are identical could not be rejected. Thus, ResNet20's test errors were deemed to not be significantly different from those of PNN15's ResNet20.

For the second pair of U tests, we considered a sample of $n_b$ observations {7.51, 7.66, 7.25, 7.48, 7.89} for WideResNet14 and a sample of $n_p$ observations {7.49, 7.26, 7.22, 7.63, 7.24} for PNN15's WideResNet14. Using SciPy's statistical function for the U test, we generated a U statistic of 19.00 with a p value = 0.22. Given α = 0.05, the null hypothesis that WideResNet14 and PNN15's WideResNet14 are identical could not be rejected. Therefore, the difference between WideResNet14 and PNN15's WideResNet14 was judged to be insignificant with respect to test errors.

**Dual Network Architecture:** The four dual network models depicted in **Fig. 6**. exhibited interesting distinctions. H and U tests were performed to measure the statistical significance of the differences between them on CIFAR-10.

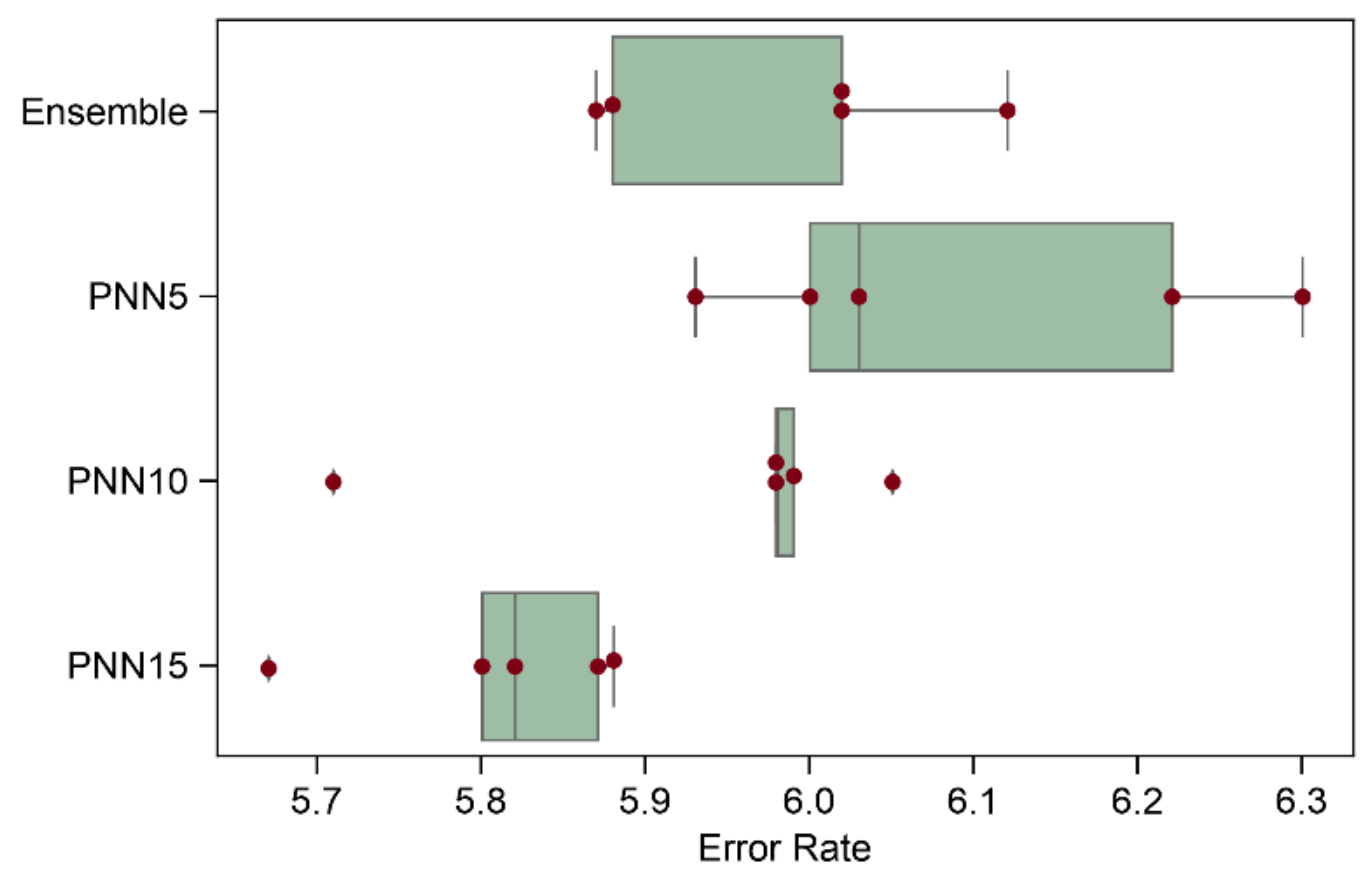

**Fig. 6.** Error Rates of Dual Network Models on CIFAR-10.

The H test hypotheses for dual network models on CIFAR-10 are listed below:

$$H_0: \tilde{x}_1 = \tilde{x}_2 = \tilde{x}_3 = \tilde{x}_4$$
$$H_1: at\ least\ one\ population\ median\ is\ differnt\ from\ those\ of\ the\ other\ models$$

Based on (1) and using SciPy's statistical function for the H test, we generated a statistic of 9.45 with a p value of 0.02. Given α = 0.05, we rejected $H_0$ to conclude that at least one population median of a model was statistically different from those of other models.

Next, we evaluated the significance of the differences between the best fine-tuned model, PNN15, and the baseline Ensemble model on CIFAR-10 using the U test. The hypotheses of the U test are listed below, where $\tilde{x}_b$ represents the population median of the baseline Ensemble (ResNet20 + WideResNet14) model and $\tilde{x}_p$ represents the population median of the proposed PNN15 (ResNet20 + WideResNet14) model.

$H_0: \tilde{x}_b = \tilde{x}_p$, two population medians are identical

$H_1: \tilde{x}_b \neq \tilde{x}_p$, two population medians are different

For the U test, we considered a sample of $n_b$ observations {6.02, 5.87, 6.02, 5.88, 6.12} for the Ensemble (ResNet20 + WideResNet14) model and a sample of $n_p$ observations {5.87, 5.80, 5.67, 5.82, 5.88} for the PNN15 (ResNet20 + WideResNet14) model. Using SciPy's statistical function of the U test, we generated a U statistic of 2.00, with a p value = 0.04. Given α = 0.05, we rejected $H_0$ and concluded that the difference between the Ensemble (ResNet20 + WideResNet14) and PNN15 (ResNet20 + WideResNet14) models was statistically significant.

*B. Discussion of Results on CIFAR-100*

**Visual Observation:** The experimental results on CIFAR-100 were plotted jointly to illustrate the error rates generated by all 12 models with both single and dual network architectures (**Fig. 7**.). The eight models with a single-network architecture exhibited a mean error rate of 23.53, whereas the four with a dual-network architecture presented a mean error rate of 20.91. Owing to the obvious error differences between the mean errors of the single and dual network models, we decided to analyze the two cases separately.

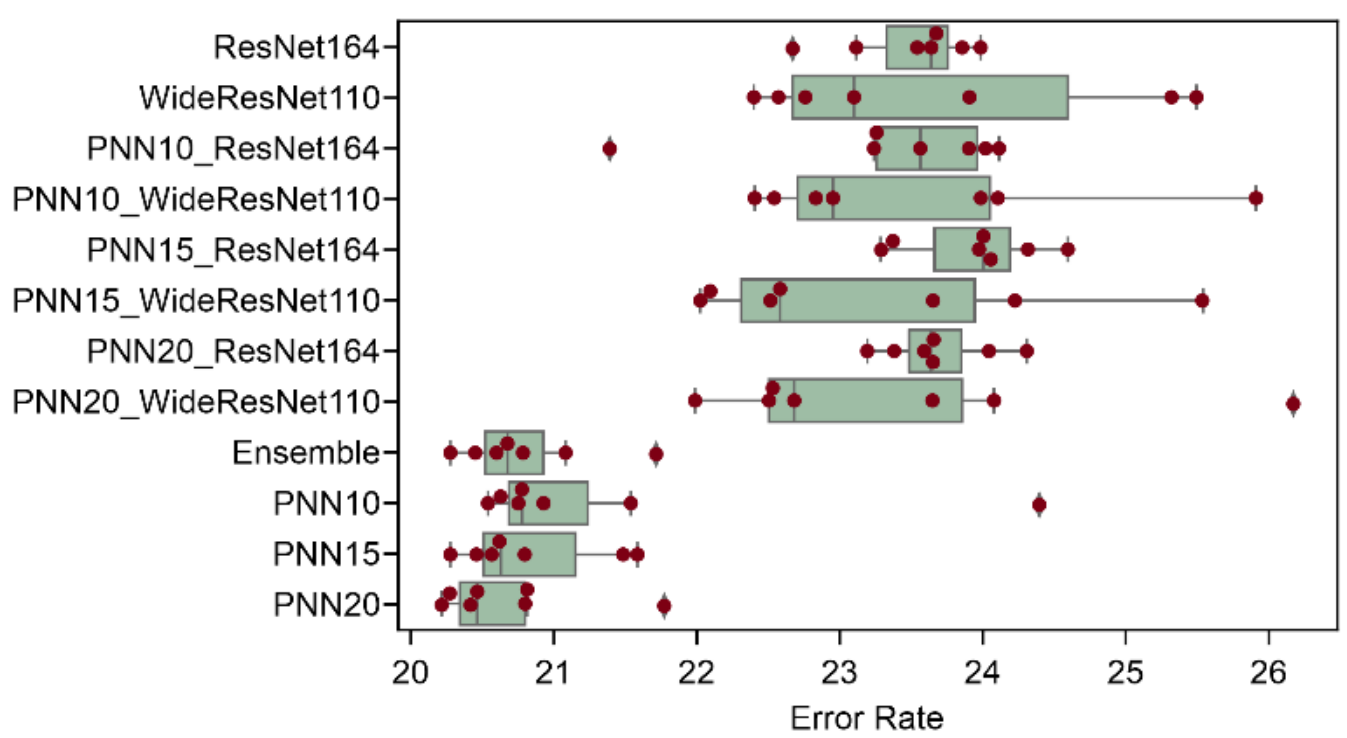

**Fig. 7**. Error rates of the models on CIFAR-100.

**Single Network Architecture:** Based on the data visualization presented in **Fig. 8**., the differences in model performance did not appear to be statistically significant. Several H and U tests were performed to confirm this conclusion.

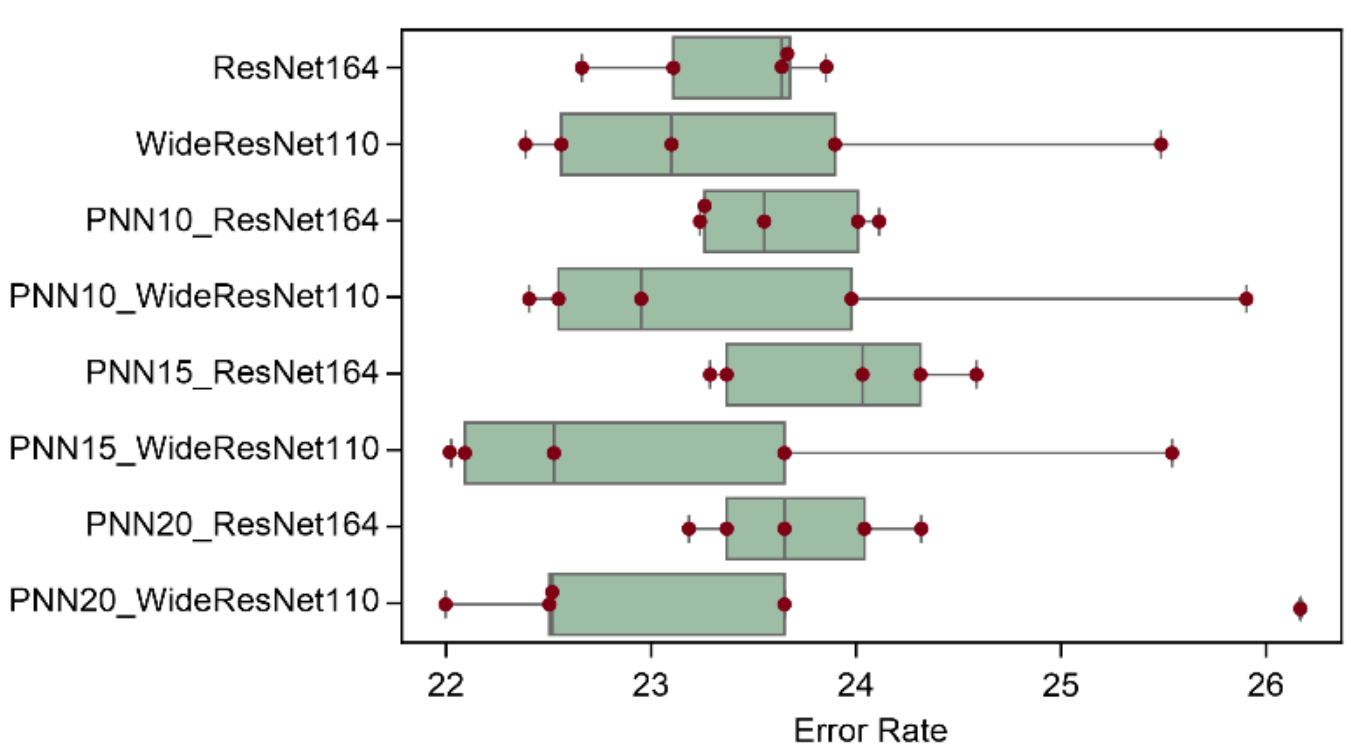

**Fig. 8.** Error rates of single-network models on CIFAR-100.

In the H test on single network models on CIFAR-100, we assumed the following:

$$H_0: \tilde{x}_1 = \tilde{x}_2 = \tilde{x}_3 = \tilde{x}_4 = \tilde{x}_5 = \tilde{x}_6 = \tilde{x}_7 = \tilde{x}_8$$

$$H_1: \textit{at least one population median is different from those of the other models}$$

The null hypothesis stated that all the single networks have identical population medians. By contrast, the alternative hypothesis claimed that at least one population median was significantly different from the others. Using SciPy's H test function in (1), we generated an H statistic of 4.82 with a p value of 0.68. Given α = 0.05, there was not sufficient evidence to reject the null hypothesis.

However, we observed a moderate performance improvement in the proposed models. As discussed in **TABLE V**, each baseline model contained three comparable PNNs. First, ResNet164 was compared with PNN10's ResNet164, PNN15's ResNet164, and PNN20's ResNet164. Among the four ResNet164 models, PNN10's ResNet164 exhibited the lowest test error mean (23.35), the second-best trimmed error mean (23.60), and the best test error median (23.56). In comparison, ResNet164 exhibited a test error mean of 23.49, a trimmed error mean of 23.56, and a test error median of 23.64. Next, WideResNet110 was compared with PNN10's WideResNet110, PNN15's WideResNet110, and PNN20's WideResNet110. Among these four WideResNet110 models, PNN15's WideResNet110 exhibited the lowest test error mean (22.23), the lowest trimmed error mean (23.01), and the lowest error median (22.58). In comparison, WideResNet110 exhibited a test error mean of 23.64, a test trimmed error mean of 23.53, and a test error median of 23.10. We conducted U tests to verify the statistical significance between ResNet164 and PNN10's ResNet164, as well as between WideResNet110 and PNN15's WideResNet110.

The following U test hypotheses were adopted:

$H_0: \tilde{x}_b = \tilde{x}_p$, two population medians are identical

$H_1: \tilde{x}_b \neq \tilde{x}_p$, two population medians are different

The null hypothesis stated that the population medians of the baseline ($\tilde{x}_b$) and proposed models ($\tilde{x}_p$) are identical on CIFAR-100. The alternative hypothesis stated that the difference between them is statistically significant.

First, we performed a U test between ResNet164 and PNN10's ResNet164. The test errors of ResNet164 were 23.64, 23.67, 23.11, 22.66, 23.85, 23.54, and 23.99, whereas those of PNN10 were 24.01, 23.26, 23.56, 23.24, 24.11, 23.91, and 21.38 (**TABLE V**). Using SciPy's U test function, we generated a U statistic of 20.00, with a p value of 0.80. Given that α = 0.05, the null hypothesis could not be rejected. We observed a moderate performance improvement in the PNN ResNet164 model; however, this improvement was not statistically significant.

Next, a U test was performed between WideResNet110 and PNN15's WideResNet110. WideResNet110's test errors were

23.10, 22.39, 25.49, 23.90, 22.56, 22.76, and 25.31, whereas those of PNN15's WideResNet110 were 22.09, 22.02, 25.54, 23.65, 22.52, 22.58, and 24.23 (**TABLE V**). Using SciPy's U test function, we generated a statistic of 30.00 with a p value of 0.52. Given α = 0.05, the null hypothesis could not be rejected. A moderate performance improvement was observed for PNN15's WideResNet110; however, this improvement was not statistically significant.

**Dual Network Architecture:** The mean test errors of the four dual network models are illustrated in **Fig. 9.** Compared with the other models, PNN20 exhibited the best mean test error, trimmed test error mean, and test error median values of 20.67, 20.55, and 20.46, respectively. We observed a trend of decreasing error rates from PNN10 to PNN15 to PNN20. First, an H test was performed to evaluate whether the differences between the four models were statistically significant. We then performed a U test to evaluate the significance of the difference between PNN20 (ResNet164 + WideResNet110) and the Ensemble (ResNet164+WideResNet110).

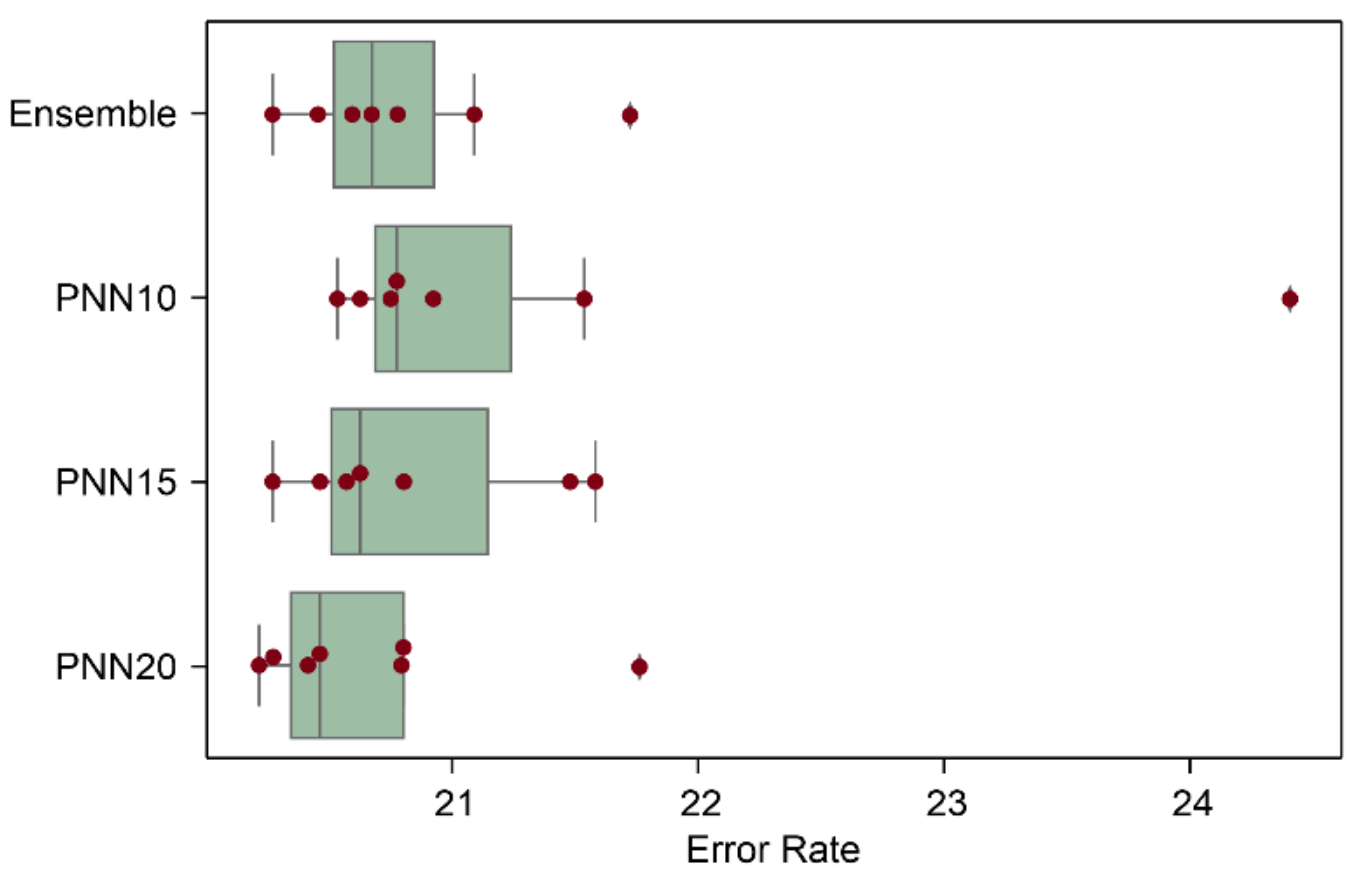

**Fig. 9.** Error Rates of Dual Network Experiments using CIFAR-100.

An H test was performed to gauge the statistical significance of the differences between the Ensemble (ResNet164+WideResNet110), PNN10 (ResNet164+WideResNet110), PNN15 (ResNet164+WideResNet110), and PNN20 (ResNet164+WideResNet110) models. The following hypotheses were tested:

$$H_0: \tilde{x}_e = \tilde{x}_{p10} = \tilde{x}_{p15} = \tilde{x}_{p20}$$

$$H_1: at\ lease\ one\ population\ median\ is\ different\ from\ those\ of\ the\ other\ models$$

Using SciPy's H test function, we generated an H statistic of 2.18, with a p value of 0.54. Given α = 0.05, there was not sufficient evidence to reject the null hypothesis. However, as discussed in **TABLE VI**, we observed a decreasing trend in the test errors of the fine-tuned PNN models—the mean test error of PNN20 (ResNet164 + WideResNet110) appeared to be different from that of the baseline Ensemble model.

Hence, we conducted a U test to further evaluate the significance of the difference between the mean test errors of the Ensemble (ResNet164+WideResNet110) and PNN20 (ResNet164+WideResNet110) models.

The following U test hypotheses were adopted:

$H_0$: $\tilde{x}_b = \tilde{x}_p$, two population medians are identical

$H_1$: $\tilde{x}_b \neq \tilde{x}_p$, two population medians are different

The null hypothesis asserted that the difference between the population medians of the baseline Ensemble (ResNet164 + WideResNet110) model ($\tilde{x}_b$) and the PNN20 (ResNet164 + WideResNet110) model ($\tilde{x}_p$) is not statistically significant on CIFAR-100. The alternative hypothesis stated that it is significantly significant.

As reported in **TABLE VI**, the Ensemble model's test errors were 20.77, 20.27, 21.08, 20.45, 20.67, 20.59, and 21.72 over the seven runs, and those of the PNN20 model were 20.27, 20.21, 21.76, 20.8, 20.46, 20.41, and 20.79. Using SciPy's H test function, we generated an H statistic of 28.50, with a p value of 0.64. Given α = 0.05, the null hypothesis could not be rejected.

We observed moderate improvements in the PNN20 model; however, these improvements were not statistically significant. The seven-run mean, trimmed mean, and median of PNN20 were all slightly better than those of the Ensemble model (**TABLE VI**).

## VI. Conclusions

In this study, we proposed a PNN inspired by the neural structures of planarians. The PNN framework consists of two embedded neural networks that communicate with each other to enhance performance compared with baseline tests. Our experiments demonstrated that the cross-communication-based PNN models, namely, PNN15 on CIFAR-10 and PNN20 on CIFAR-100, outperformed the other models. This was corroborated by the lower mean test errors of the former model than those of the latter on both datasets, with the difference being statistically significant on CIFAR-10. Overall, this study highlighted the effectiveness of the PNN framework in improving the performance of neural networks in image classification tasks.

## VII. Future Works

PNNs are innovative tools that can be used to construct ANNs inspired by the biological evolutionary patterns of lower and higher life forms. The PNN proposed in this study mimicked the biological nervous systems of planarians. It was constructed by combining two ResNet networks of varying depths and widths into a single model, in which the networks were connected by an artificial brain.

A PNN enables the integration of several neural networks to collaborate and generate enhanced optimal results. Future improvements to PNNs may include (1) more densely connected weight-swapping mechanisms, (2) enhanced brain and nerve cord interactions, (3) further optimization of global epochs and subepochs, (4) possible changes inspired by other types of biological nervous systems, and (5) potential integration with other types of artificial neural networks, such as large language models, to extend the cross-communication capabilities of the PNN framework, enabling it to emulate human-like cognitive processes and enhance human–machine interactions.

ACKNOWLEDGMENT

The authors would like to extend their sincere gratitude to Dr. Yunzhan Li, Department of Cellular and Molecular Physiology, Penn State College of Medicine, Hershey, Pennsylvania, USA, for his invaluable insights into biological neural networks and developmental biology. His guidance and constructive feedback greatly contributed to the progress of this work on *Planarian Neural Networks: Evolutionary Patterns from Basic Bilateria Shaping Modern Artificial Neural Network Architectures* and were instrumental in shaping the outcome of this study.